\documentclass{article}

\usepackage[preprint]{neurips_2026}

\usepackage[utf8]{inputenc}
\usepackage[T1]{fontenc}
\usepackage[table,dvipsnames]{xcolor}
\definecolor{NavyBlue}{RGB}{0,0,128}
\usepackage[colorlinks=true,linkcolor=NavyBlue,citecolor=NavyBlue,urlcolor=NavyBlue]{hyperref}
\usepackage{url}
\usepackage{graphicx}
\usepackage{wrapfig}
\usepackage{booktabs}
\usepackage{multirow}
\usepackage{amsmath}
\usepackage{amssymb}
\usepackage{amsfonts}
\usepackage{mathtools}
\usepackage{bm}
\usepackage{nicefrac}
\usepackage{microtype}
\usepackage{tikz}
\usepackage{colortbl}
\usepackage{algorithm}
\usepackage{algpseudocode}
\usepackage{enumitem}
\newcommand*\circledblue[1]{\tikz[baseline=(char.base)]{
            \node[shape=circle,draw=NavyBlue!100,fill=NavyBlue!10,thick,inner sep=1pt] (char) {\scriptsize\textsf#1};}}

\newcommand{\rqtag}[2]{
\tikz[baseline=(rq.base)]\node[
    inner sep=2.0pt,
    rounded corners=2.5pt,
    draw=#1,
    line width=0.8pt,
    fill=#1!12,
    text=#1!85!black,
    font=\bfseries\footnotesize
](rq){RQ#2};
}
\definecolor{rq1}{HTML}{1F77B4}
\definecolor{rq2}{HTML}{FF7F0E}
\definecolor{rq3}{HTML}{2CA02C}
\definecolor{rq4}{HTML}{D62728}
\definecolor{rq5}{HTML}{9467BD}
\definecolor{rq6}{HTML}{8C564B}
\definecolor{rq7}{HTML}{17BECF}
\definecolor{rq8}{HTML}{BCBD22}
\definecolor{rq9}{HTML}{7F7F7F}
\definecolor{CentralFail}{RGB}{235,242,250}
\definecolor{BoundaryFail}{RGB}{255,242,230}
\definecolor{NearSuccess}{RGB}{236,248,240}
\definecolor{SoftPurple}{HTML}{B8A8D8}
\setcitestyle{round}

\title{\textsc{SkillGen}: Verified Inference-Time Agent Skill Synthesis}

\author{
\begin{tabular}{@{}l@{\hspace{1.2em}}l@{\hspace{1.2em}}l@{\hspace{1.2em}}l@{}}
{\bfseries Yuchen Ma\textsuperscript{1}\thanks{Equal contribution. Contact emails: \texttt{yuchen.ma@lmu.de} and \texttt{yhuang37@nd.edu}.}} &
{\bfseries Yue Huang\textsuperscript{2}\footnotemark[1]} &
{\bfseries Han Bao\textsuperscript{2}} &
{\bfseries Haomin Zhuang\textsuperscript{2}} \\
{\bfseries Swadheen Shukla\textsuperscript{3}} &
{\bfseries Michel Galley\textsuperscript{3}} &
{\bfseries Xiangliang Zhang\textsuperscript{2}} &
{\bfseries Stefan Feuerriegel\textsuperscript{1}}
\end{tabular} \\
\textsuperscript{1}Munich Center for Machine Learning, LMU Munich \\
\textsuperscript{2}University of Notre Dame \quad
\textsuperscript{3}Microsoft Research
}

\begin{document}

\maketitle

\begin{abstract}

Skills are a promising way to improve LLM agent capabilities without retraining, while keeping the added procedure reusable and controllable. However, high-quality skills are still largely written by hand. We introduce \textsc{SkillGen}, a multi-agent framework that synthesizes a single auditable skill from trajectories generated by a base agent. The output is a human-readable artifact that can be inspected before use. Rather than merely summarizing trajectories, \textsc{SkillGen} leverages contrastive induction over both successful and failed trajectories to identify reusable success patterns, recurring failure modes, and behaviors that appear in nearby successes but are missing from failures. \textsc{SkillGen} then generates candidate skills and iteratively refines the skill. A key novelty in \textsc{SkillGen} is that we model agent skills as interventions to empirically verify the net effect of skills on the overall performance. Specifically, we compare outcomes on the same instances with and without the skill, so that we account for both repairs (cases where the skill fixes a baseline failure) and regressions (cases where the skill breaks a baseline success). Across a broad range of agents and datasets, \textsc{SkillGen} consistently improves held-out performance, outperforms existing skill-generation baselines, and produces skills that transfer across models.

\end{abstract}

\section{Introduction}
\label{sec:introduction}

Large language models (LLMs) are increasingly used to solve complex, multi-step tasks \citep{schick2023toolformer,qin2024toolllm,yao2023react,wang2023voyager}. A common way to formalize such behavior is through \textit{\textbf{skills}}: reusable, inference-time procedures that encode task-specific guidance, such as instructions, executable code, and domain knowledge, without modifying model weights \citep{zhang2025agentskills,anthropic2025skills}. Skills are modular and auditable: because they are
readable inference-time artifacts rather than weight updates or
prompt searches, one can inspect the procedure they encode, revise it directly, and test its effect before deployment. In practice, however, high-quality skills are still largely hand-written.

Automated skill synthesis aims to learn reusable skills from agent experience \citep{shinn2023reflexion,zhao2024expel,ni2026trace2skill,alzubi2026evoskill,wang2026skillx,zhang2026evoskills}. However, existing methods have two key {shortcomings}. First, existing methods primarily learn from successful trajectories, and even when failures are considered, they are typically summarized in isolation rather than contrasted against nearby successes on the same task. As a result, prior work misses a key contrastive signal between success and failure---that is, what the agent executes correctly in similar contexts and what it omits in failed roll-outs. For example, a successful trajectory may include an intermediate validation step that is absent in a failed attempt, but success-only learning does not isolate that such intermediate validation is important and would not put into a reusable pattern. Second, existing methods do not explicitly \emph{verify} the empirical benefit of a generated skill. While a skill may repair some failures, it can also introduce new failure modes on cases that the agent previously solved correctly. As a result, skill synthesis is fundamentally an interventional problem, where one compares the net-effect on the agent's performance with and without the candidate skill. Also, such performance evaluation is necessary to eventually build iterative approaches to refine candidate skills in a principled manner.

\begin{figure*}[t]
\centering
\includegraphics[width=1\textwidth]{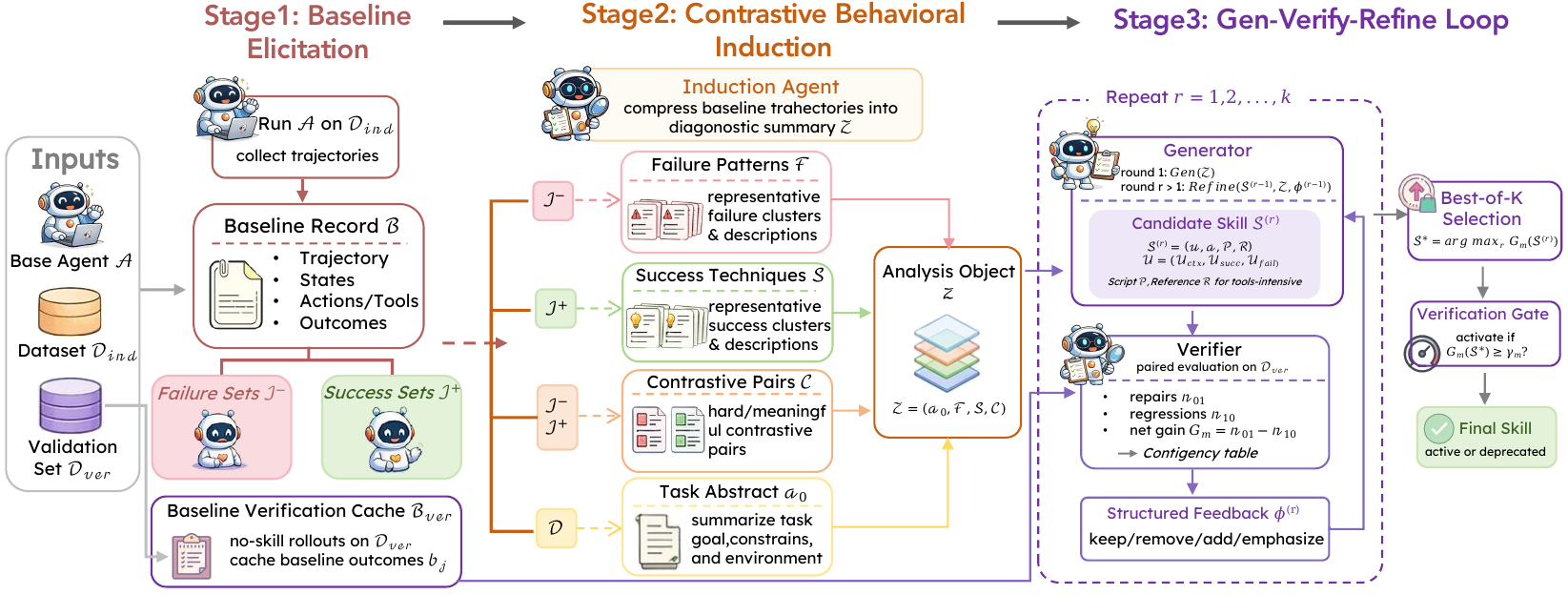}
\vspace{-12pt}
\caption{\textbf{\textsc{SkillGen} overview.} Our multi-agent framework synthesizes a single auditable skill from baseline trajectories. \protect\circledblue{1} It first elicits successful and failed rollouts as input. \protect\circledblue{2} It extracts reusable patterns of successful and failure modes. \protect\circledblue{3} It follows an iterative generation-verification-refinement loop to generate and refine new candidate skills.}
\label{fig:overview}
\vspace{-12pt}
\end{figure*}

We introduce \textsc{SkillGen}: a \emph{multi-agent framework for automatic, inference-time skill synthesis} (see Fig.~\ref{fig:overview}). \textsc{SkillGen} takes an existing dataset of LLM trajectories as input and derives a single auditable skill: a readable intervention whose task context, success procedures, and failure lessons can be inspected and whose empirical net effect is verified before deployment. The input dataset can be collected during a baseline elicitation phase to compile successful and failed trajectories.  Our framework operates through three specialized agents: (1)~A \textbf{contrastive induction agent} analyzes the input trajectories to extract reusable success patterns \underline{and} identify recurring failure modes, with the aim to surface contrasts between successful vs. failed roll-outs. As a result, it outputs a compact and interpretable summary with task diagnostics. (2)~In a generation-verification-refinement loop, the diagnostics are converted into candidate skills, and the skills are then iteratively refined based on feedback (using a  \textbf{generation agent} and a \textbf{verification agent}. The final skill is selected by measuring the net-effect on the final held-out performance. This ensures that the selected skill improves the overall performance and thus accounts for \textit{``repairs''} (i.e, when a skill fixes a failure) and \textit{``regressions''} (i.e., when a skill breaks a correct case). To the best of our knowledge, \textsc{SkillGen} is the first agentic framework that models inference-time skill synthesis as an intervention problem to ensure a positive, empirically verified effect on performance.

We also demonstrate the effectiveness of \textsc{SkillGen} across a broad range of interactive, scientific, coding, and other tool-use benchmarks. We further evaluate \textsc{SkillGen} using several open-weight and proprietary base LLMs. As our main result, \textsc{SkillGen} improves average accuracy for all eight evaluated base LLMs, with held-out gains ranging from $+3.27$ to $+10.08$ percentage points. We also compare \textsc{SkillGen} against state-of-the-art skill-generation baselines~\citep{ni2026trace2skill,wang2026skillx,alzubi2026evoskill,zhang2026evoskills}, where \textsc{SkillGen} is consistently positive and achieves the largest average improvement by a clear margin. Our ablations show that contrastive induction, verification-guided refinement, and the verification gate each contribute to the performance gains. We also perform a cross-model transfer analysis to show that generated skills are generalizable and not tied to the LLM that produced them.

\noindent\textbf{Contributions.}
Our main contributions are three-fold:\footnote{Code is available via \url{https://github.com/yccm/SkillGen}. }{(1)}~We formulate a general, end-to-end learning task for automatic inference-time skill synthesis: to produce a single, auditable skill that improves a base agent. {(2)}~We introduce \textsc{SkillGen}, a multi-agent framework that learns from both failed and successful trajectories via contrastive induction, and then generates new candidate skills that are iteratively refined and verified. The final skills are selected to have a positive net-effect on the overall performance. {(3)}~We provide an extensive empirical study to demonstrate consistent and large held-out performance gains. \textsc{SkillGen} outperforms state-of-the-art skill-generation baselines and produces skills that transfer across models without parameter updates.

\section{Preliminaries}
\label{sec:prelim}

We view inference-time skills as \emph{interventions} that modify the behavior of a base agent and thereby change its task performance. This perspective naturally induces a comparison between outcomes with and without a given skill on the same inputs.

\textbf{Task setting.} Let $\mathcal{X}$ be the input space, $p$ a task distribution over $\mathcal{X}$, and $\mathcal{T}$ the space of agent trajectories. A trajectory $\tau \in \mathcal{T}$ consists of the full sequence of LLM interactions, including messages, tool calls, environment observations, and the final output. For skill synthesis, we split the training data into: (i) an induction subset $\mathcal{D}_{\mathrm{ind}}=\{x_i\}_{i=1}^{n}$ used to analyze agent behavior, and (ii) a construction-time verification subset $\mathcal{D}_{\mathrm{ver}}=\{\tilde{x}_j\}_{j=1}^{m}$ used for evaluating and selecting candidate skills. We consider a \emph{base agent} $\mathcal{A}$ that maps inputs to trajectories and that we seek to improve upon. We model $\mathcal{A}$ as a stochastic trajectory kernel $P_{\mathcal{A}}(\tau\mid x;\eta)$,
where $x$ is the task instance and $\eta$ is an inference-time intervention loaded into the agent's context.
The empty intervention $\eta=\varnothing$ defines the ``no-skill'' behavior, defined by $\tau^{0}(x)\sim P_{\mathcal{A}}(\cdot\mid x;\varnothing)$.

 To formalize the outcome $Y$, we define a task-level evaluator $\mathcal{E}:\mathcal{X}\times\mathcal{T}\!\rightarrow\![0,1]$. In practice, this could be an LLM-as-a-judge, a benchmark score, or a successful check against some environment outcome. As a result, the evaluator assigns a success probability to each instance--trajectory pair. The observed outcome is $Y(x,\tau)\sim\operatorname{Bernoulli}(\mathcal{E}(x,\tau))$, with deterministic evaluators as the special case $\mathcal{E}(x,\tau)\in\{0,1\}$. For any instance $x$, we define the baseline outcome $Y^{0}(x)=Y(x,\tau^{0}(x))$; for induction instances, we write $\tau_i^{0}=\tau^{0}(x_i)$, and let $y_i^{0}$ denote the realized outcome of the base agent.

\textbf{Skill interventions:} We define a candidate \emph{skill} as a
inference-time intervention $s=(u,a,\mathcal{P},\mathcal{R}),$
where $u$ is a structured prompt, $a$ is task metadata (e.g., task description), $\mathcal{P}$ is an optional set of executable scripts, and $\mathcal{R}$ is an optional collection of auxiliary documents.
Together, these components define the skill space considered by \textsc{SkillGen}.

We model skills as interventions that change the agent's behavior and thus its outcomes. To make comparative assessments of skill learning, we adopt the potential outcome framework \citep{rubin2005causal} as a principled manner to formalize the treatment effects. For any input $x$ and a candidate skill $s$, we define two potential outcomes: the baseline outcome (i.e., $Y^{0}(x)=Y(x,\tau^{0}(x))$) and the skill-augmented outcome (i.e., $Y^{s}(x)=Y(x,\tau^{s}(x))$, $\tau^{s}(x)\sim P_{\mathcal{A}}(\cdot\mid x;\eta(s))$.
Loading a skill $s$ corresponds to applying an intervention $\eta(s)$ into $\mathcal{A}$.

\textbf{Objective:}
Our goal is to measure and maximize the expected effect of a skill relative to the baseline agent:
\begin{equation}
\label{eq:delta}
    \Delta(s)
    \;=\;
    \mathbb{E}_{x\sim p}
    \left[
        \mathbb{E}\!\left[Y^{s}(x)\mid x,s\right]
        -
        \mathbb{E}\!\left[Y^{0}(x)\mid x\right]
    \right].
\end{equation}
Thus, $\Delta(s)$ captures the net-effect induced by a skill intervention $s$: it measures how much the skill improves (or degrades) performance on the same input distribution, while accounting for both ``repairs'' (i.e., cases where the skill fixes a baseline failure) and ``regressions'' (i.e., cases where the skill breaks a baseline success). The objective for skill synthesis is therefore to select a skill with positive net-effect on held-out performance, but without relying on human-written task-specific skills. During construction, each candidate skill is evaluated on $\mathcal{D}_{\mathrm{ver}}$ under identical inputs with and without the skill. As a result, we yield a so-called \textit{status} $\sigma_{\mathrm{ver}}(s;\mathcal{D}_{\mathrm{ver}})\in\{\mathrm{active},\mathrm{deprecated}\}$.
At deployment, only active skills are loaded; deprecated skills are subsumed under the empty intervention $\varnothing$.

\section{\textsc{SkillGen}}
\label{sec:method}

\textbf{Overview.} \textsc{SkillGen} takes as input: a base agent, a set of observed LLM trajectories (split into an induction subset and verification subset), and a task-level evaluator. \textsc{SkillGen} then returns a single auditable skill.
\textsc{SkillGen} follows an agentic, three-staged framework (see Fig.~\ref{fig:overview}; pseudocode in Alg.~\ref{alg:skillgen-loop} in Appendix~\ref{app:algorithm}).
\emph{Stage}~\circledblue{1} baseline elicitation: this stage uses the base agent to collect successful vs. failed trajectories.
\emph{Stage}~\circledblue{2} contrastive induction: this stage extracts recurring failure modes to identify local patterns that distinguish successful vs. failed roll-outs; the patterns are combined into a compact, interpretable summary of task-level diagnostics (via the \textbf{induction agent}).
\emph{Stage}~\circledblue{3} an iterative generation--verification--refinement loop: this stage turns the diagnostics into candidate skills  (via the \textbf{generation agent}), tests each candidate skill on the verification subset for performance evaluation (via the \textbf{verification agent}), and finally makes refinements to the candidate skill.
Finally, the candidate skill with the largest construction-time net-effect $\Delta(s)$ is returned.

\subsection{\emph{Stage}~\texorpdfstring{\protect\circledblue{1}} {(1)} : Baseline Elicitation}

We first run the base agent on the induction subset and store
\begin{equation}
    \mathcal{B}=\{(x_i,\tau^{0}_{i},y^{0}_{i})\}_{i=1}^{n},
    \qquad
    \mathcal{I}^{-}=\{i:y^{0}_{i}=0\},\quad
    \mathcal{I}^{+}=\{i:y^{0}_{i}=1\}.
\end{equation}
Here, $\mathcal{I}^{-}$ indexes failed baseline rollouts and $\mathcal{I}^{+}$ indexes successful ones.
Intuitively, failures show where the base agent would need help; and successes show procedures the base agent can already execute.
Using both strata is unique to \textsc{SkillGen} and important: failures alone can produce misleading or unhelpful advice, while successes alone do not identify the capability gap.

We further cache no-skill outcomes on the construction-time verification subset, i.e.,
\begin{equation}
    \mathcal{B}_{\mathrm{ver}}
    =
    \{(\tilde{x}_j,\tilde{\tau}^{0}_{j},b_j)\}_{j=1}^{m},
    \qquad
    \tilde{\tau}^{0}_{j}\sim P_{\mathcal{A}}(\cdot\mid \tilde{x}_j;\varnothing),
    \qquad
    b_j=Y(\tilde{x}_j,\tilde{\tau}^{0}_{j}).
\end{equation}
The cached outcomes are neither used by the induction agent nor the first-round generation agent; instead, the cached outcomes are later used for construction-time verification and subsequent refinement. The main motivation is that we can later compare each candidate's skill against the behavior of the no-skill agent on the same verification subset.

\subsection{\emph{Stage}~\texorpdfstring{\protect\circledblue{2}}{(2)}: Contrastive Behavioral Induction}

The \textbf{induction agent} compresses baseline trajectories into an explicit summary diagnostic for skill synthesis:
\begin{equation}
    \operatorname{Compress}(\mathcal{B})
    =
    \mathcal{Z}=(a_0,\mathcal{F},\mathcal{S},\mathcal{C}),
\end{equation}
where $a_0$ is a task-level summary of the induction inputs, $\mathcal{F}$ is a set of cluster-level failure summaries, $\mathcal{S}$ is a set of cluster-level success summaries, and $\mathcal{C}$ is a set of local contrastive observations between nearby failed and successful rollouts. Any of the three set-valued components may be empty (e.g., if the corresponding success or failure stratum is empty. Stage~\circledblue{3} receives only $\mathcal{Z}$, so this stage converts variable-length LLM trajectories into a lower-dimensional diagnostic summary for skill generation.

\noindent$\bullet$\,\textbf{Task summary ($a_0$).}
The induction agent applies a fixed abstraction prompt, denoted by $\operatorname{Abs}$, to the induction inputs and writes a task summary $a_0=\operatorname{Abs}(\{x_i\}_{i=1}^{n})$. The summary is designed to describe the task family rather than any single instance, giving the generation agent a general description of what the skill is for. The component $a_0$ is part of the diagnostic summary $\mathcal{Z}$ and is distinct from the skill metadata $a$ in $s=(u,a,\mathcal{P},\mathcal{R})$.

\noindent$\bullet$\,\textbf{Failure analysis ($\mathcal{F}$).}
For each failed rollout $i\in\mathcal{I}^{-}$, the induction agent writes a root-cause summary $\rho_i^{-}$.
Let $\phi$ be a text encoder applied to a serialized input--summary pair, and define $e_i^{-}=\phi([x_i;\rho_i^{-}])$.
We cluster the resulting failure embeddings via
\begin{equation}
    \Pi^{-}=\operatorname{Cluster}\big(\{e_i^{-}:i\in\mathcal{I}^{-}\}\big),
    \qquad
    f_P=\operatorname{Summ}^{-}\big(\{(x_i,\tau_i^{0},\rho_i^{-}):i\in P\}\big)
\end{equation}
for each cluster $P\in\Pi^{-}$.
Each $f_P$ is a cluster-level failure summary: it describes the recurring root cause, the trajectory point at which the failure typically appears, and the corrective rule that would avoid the failure.
The resulting set is $\mathcal{F}=\{f_P\}_{P\in\Pi^{-}}$.

\noindent$\bullet$\,\textbf{Success analysis ($\mathcal{S}$).}
For each $i\in\mathcal{I}^{+}$, an LLM writes a success summary $\rho_i^{+}$, and we further embed $e_i^{+}=\phi([x_i;\rho_i^{+}])$.
We apply the same embedding and clustering procedure to successful rollouts:
\begin{equation}
    \Pi^{+}=\operatorname{Cluster}\big(\{e_i^{+}:i\in\mathcal{I}^{+}\}\big),
    \qquad
    h_P=\operatorname{Summ}^{+}\big(\{(x_i,\tau_i^{0},\rho_i^{+}):i\in P\}\big),
\end{equation}
and $\mathcal{S}=\{h_P\}_{P\in\Pi^{+}}$.
Each $h_P$ is a cluster-level success summary: it describes the reusable procedure, the task conditions under which it appears, and checks that make the procedure robust.

\noindent$\bullet$\,\textbf{Local contrastive analysis ($\mathcal{C}$).}
The above cluster summaries can miss some of the ``small'' action choices that separate a success from a failure.
When $\mathcal{I}^{+}$ is non-empty, for each failed instance $i\in\mathcal{I}^{-}$, we retrieve the nearest successful neighbor under embedding distance $d$, i.e.,
\begin{equation}
    j(i)=\operatorname*{arg\,min}_{j\in\mathcal{I}^{+}}\,d(e_i^{-},e_j^{+}).
\end{equation}
The induction agent first checks whether $x_i$ and $x_{j(i)}$ share the same task type.
If they do, the induction agent compares the two full trajectories and generates a contrastive observation
\[
    c_i=\operatorname{Contr}(x_i,\tau_i^{0},x_{j(i)},\tau_{j(i)}^{0}),
\]
which describes the behavior present in the successful rollout and the corresponding behavior omitted in the failed rollout. The observations whose pairs pass the same-task check form $\mathcal{C}$.
Thus, $\mathcal{C}$ provides local contrastive evidence: it anchors advice in behavior that the same base agent has already demonstrated, but that was absent in a nearby failure.

\subsection{\emph{Stage}~\texorpdfstring{\protect\circledblue{3}}{(3)}: Generation--Verification--Refinement Loop}
\label{sec:paired-verification}

\textbf{Overview.}
Stage~\circledblue{3} turns the diagnostic summary $\mathcal{Z}$ into a sequence of candidate skills and uses paired verification to decide which candidate should be deployed.
The loop is designed to improve baseline failures while explicitly tracking regressions on instances that the base agent already solved.
It consists of four steps:
\textbf{(i)}~\emph{generation}, which produces candidate skills from the diagnostic summary;
\textbf{(ii)}~\emph{verification}, which evaluates each candidate on the verification subset;
\textbf{(iii)}~\emph{refinement}, which updates candidates using structured feedback from repairs and regressions; and
\textbf{(iv)}~\emph{selection}, which returns the candidate with the largest verified net gain for deployment.
Let $K\ge1$ denote the round budget. We index rounds by $r\in\{1,\ldots,K\}$, write $s^{(r)}$ for the candidate skill at round $r$, write $\Phi^{(r)}$ for the feedback produced after verifying $s^{(r)}$, and write $s^\star$ for the selected skill.

$\bullet$\,\underline{\textbf{(i) Generation:}} In each round $r$, the \textbf{generation agent} uses the diagnostic summary $\mathcal{Z}$ as well as feedback from the previous round (with $\Phi^{(0)}=\varnothing$) to produce a new candidate skill
\begin{equation}
    s^{(r)}=(u^{(r)},a^{(r)},\mathcal{P}^{(r)},\mathcal{R}^{(r)}).
\end{equation}
\emph{Skill structure:} To write the new candidate skill, we use a prompt template with a fixed three-part schema
\begin{equation}
    u^{(r)}=(u^{(r)}_{\mathrm{ctx}},u^{(r)}_{\mathrm{succ}},u^{(r)}_{\mathrm{fail}})
\end{equation}
in natural language, where:
(i)~$u_{\mathrm{ctx}}$ encodes task context, i.e., a concise description of the task distribution and constraints (derived from $a_0$);
(ii)~$u_{\mathrm{succ}}$ encodes reusable success patterns distilled from $\mathcal{S}$ and the successful instances from the contrastive analysis $\mathcal{C}$; and
(iii)~$u_{\mathrm{fail}}$ encodes reusable failure-avoidance patterns derived from $\mathcal{F}$ and the negative instances from the contrastive analysis $\mathcal{C}$.
The above schema acts as a constrained projection from the diagnostic summary to the skill space. Intuitively, the idea here is to learn patterns with reusable procedures that define successful vs failure instances, so that the refinement can help encourage the former and avoid the latter.

For tool-intensive tasks, the generation agent may additionally emit scripts $\mathcal{P}^{(r)}$ and reference documents $\mathcal{R}^{(r)}$; however, after round $r > 1$, refinement edits are restricted to the body $u^{(r)}$ in natural language, which keeps the tool interface fixed to prevent uncontrolled expansion.

$\bullet$\,\underline{\textbf{(ii) Verification:}} The \textbf{verification agent} evaluates each candidate skill on all instances in the verification subset $\mathcal{D}_{\mathrm{ver}}$.
For a candidate skill $s$, we load the intervention $\eta(s)$ into the base agent and roll it out on each $\tilde{x}_j\in\mathcal{D}_{\mathrm{ver}}$, i.e.,
\begin{equation}
    z_j(s)=Y(\tilde{x}_j,\tilde{\tau}^{s}_{j}),
    \qquad
    \tilde{\tau}^{s}_{j}\sim P_{\mathcal{A}}(\cdot\mid \tilde{x}_j;\eta(s)).
\end{equation}
\emph{Causal evaluation of skill intervention :}
We treat a candidate skill $s$ as an intervention on the base agent and evaluate the effect by comparing outcomes with and without the intervention on the same inputs.
For each $\tilde{x}_j\in\mathcal{D}_{\mathrm{ver}}$, we observe the baseline outcome $b_j = Y^{0}(\tilde{x}_j)$ and the skill-augmented outcome $z_j(s)=Y^{s}(\tilde{x}_j)$.
Applying the skill to all verification instances yields a direct comparison between $Y^{0}$ and $Y^{s}$ on identical inputs.
In this view, \emph{``repairs''} correspond to $Y^{0}=0 \rightarrow Y^{s}=1$, while \emph{``regressions''} correspond to $Y^{0}=1 \rightarrow Y^{s}=0$.

\emph{Comparative metrics.}
We aggregate outcomes via
\begin{equation}
    n_{\alpha\beta}(s)=\sum_{j=1}^{m}\mathbf{1}\{Y^{0}(\tilde{x}_j)=\alpha,\,Y^{s}(\tilde{x}_j)=\beta\},
\end{equation}
with repairs $n_{01}(s)$ and regressions $n_{10}(s)$.
The empirical net-effect under this comparison is $\widehat{\Delta}_m(s)$,
\begin{equation}
\label{eq:netgain}
    \widehat{\Delta}_m(s)
    =\frac{1}{m}\sum_{j=1}^{m}\big(Y^{s}(\tilde{x}_j)-Y^{0}(\tilde{x}_j)\big)
    =\frac{n_{01}(s)-n_{10}(s)}{m},
    \qquad
    G_m(s)=n_{01}(s)-n_{10}(s).
\end{equation}
For a fixed, non-adaptively chosen skill and i.i.d.\ verification instances,
$\mathbb{E}[\widehat{\Delta}_m(s)]=\Delta(s)$.

$\bullet$\,\underline{\textbf{(iii)~Refinement:}} Refinement uses structured feedback to update the skill.
$\bullet$\,\emph{Feedback signals.}
After each round, the verification agent summarizes the diagnostic evidence rather than sending raw trajectories back to the generation agent. For this, the verification agent partitions instances into
\begin{equation}
\resizebox{0.98\columnwidth}{!}{$\displaystyle
\mathcal{Q}^{(r)}_{\mathrm{repair}}=\{j:b_j=0,\,z_j(s^{(r)})=1\},\quad
\mathcal{Q}^{(r)}_{\mathrm{regress}}=\{j:b_j=1,\,z_j(s^{(r)})=0\},\quad
\mathcal{Q}^{(r)}_{\mathrm{fail}}=\{j:b_j=0,\,z_j(s^{(r)})=0\},
$}
\end{equation}
Here, $\mathcal{Q}^{(r)}_{\mathrm{repair}}$ contains baseline failures repaired by $s^{(r)}$, $\mathcal{Q}^{(r)}_{\mathrm{regress}}$ contains baseline successes broken by $s^{(r)}$, and $\mathcal{Q}^{(r)}_{\mathrm{fail}}$ contains baseline failures that remain unresolved.
$\bullet$\,\emph{Feedback aggregation.}
The verification agent creates explanations of how the skill affected selected repairs, regressions, and unresolved failures. The verification agent then aggregates these explanations into
\begin{equation}
    \Phi^{(r)}=
    \big(\Phi^{(r)}_{\mathrm{keep}},
    \Phi^{(r)}_{\mathrm{remove}},
    \Phi^{(r)}_{\mathrm{add}},
    \Phi^{(r)}_{\mathrm{emphasize}}\big),
\end{equation}
which specifies, for the next round, which parts of the current skill to keep, remove, add, and emphasize.
$\bullet$\,\emph{Update rule.} The refinement uses the following update (to avoid writing a new prompt from scratch):
\begin{equation}
\label{eq:refine}
    s^{(r)}=
    \begin{cases}
        \operatorname{Gen}(\mathcal{Z}), & r=1,\\
        \operatorname{Refine}(s^{(r-1)},\mathcal{Z},\Phi^{(r-1)}), & r>1.
    \end{cases}
\end{equation}

$\bullet$\,\underline{\textbf{(iv)~Final skill selection:}} Since later refinement rounds need not improve empirical performance, \textsc{SkillGen} performs a best-of-$K$ selection over the candidate sequence $\{s^{(r)}\}_{r=1}^{K}$ and returns the candidate skill with the largest construction-time net gain $G_m$, i.e.,
\begin{equation}
    r^{\star}=\operatorname*{arg\,max}_{1\le r\le K}G_m(s^{(r)}),
    \qquad
    s^\star=s^{(r^\star)}.
\end{equation}

\emph{Verification gate.} A candidate skill is marked `active' only if it satisfies
\begin{equation}
\label{eq:gate}
    G_m(s^\star)\ge\gamma_m,
    \qquad
    \gamma_m=\max\{g_{\mathrm{abs}},\lceil g_{\mathrm{rel}}m\rceil,1\}.
\end{equation}
Otherwise, it is marked `deprecated' and replaced by the empty intervention.\footnote{Here, $g_{\mathrm{abs}}\in\mathbb{Z}_{\ge 0}$ is an absolute minimum number of net repairs, and $g_{\mathrm{rel}}\in[0,1]$ is a relative minimum as a fraction of the construction-time verification subset. The gate is a simple construction-time safeguard: the absolute term prevents deploying candidates whose gain is negligible in count, the relative term requires the gain to scale with the size of the verification subset, and the final lower bound of $1$ requires a strictly positive construction-time net gain.} The threshold due to $\gamma_m$ defines the construction-time deployment rule used by \textsc{SkillGen}: across refinement rounds, we first select the candidate with the largest $G_m$, and then mark the selected skill as active if it satisfies Eq.~\eqref{eq:gate}; otherwise, the skill is deprecated.

\emph{Deployment:} At runtime, an active skill is injected into a dedicated slot of the system prompt.
Reference documents are retrieved via on-demand loading through \texttt{skill\_load\_reference}; executable scripts encode only declared top-level functions with prefixed \texttt{skill\_}.
This ensures that the deployed capability matches the verified skill.

\section{Experiments}
\label{sec:experiments}

We evaluate \textsc{SkillGen} on held-out test instances across interactive, scientific, coding, web, and tool-use benchmarks; full implementation details are in Appendix~\ref{app:experimental-details}.
All claims use paired held-out evaluations: after construction is complete, the same task instances are rolled out with and without the generated skill.

\noindent\textbf{\rqtag{rq1}{1} Does \textsc{SkillGen} improve base agents across model families and benchmark domains?}

Before any held-out rollout, each skill and its active/deprecated status is fixed using only the skill-training dataset: the induction subset for trajectory analysis and the construction-time verification subset for refinement and selection.
Table~\ref{tab:main-results} reports the no-skill baseline accuracy, the skill-augmented accuracy, and the absolute accuracy change over 80 held-out benchmark--split--model combinations.
\begin{table*}[t]
\centering
\caption{\textbf{Main results across open-weight and proprietary models.} For each model, we report the no-skill baseline accuracy (\textsc{Base}), the skill-augmented accuracy (\textsc{Skill}), and the absolute accuracy change ($\Delta$) on held-out test instances. Values are from the paired rollout per instance under the split protocol in Appendix~\ref{app:datasets-splits}.}
\label{tab:main-results}
\vspace{3pt}
\begingroup
\footnotesize
\definecolor{sgPanel}{RGB}{232,241,250}
\definecolor{sgAvg}{RGB}{255,253,244}
\definecolor{sgStripe}{RGB}{250,251,253}
\definecolor{sgUpLow}{RGB}{244,252,247}
\definecolor{sgUpMid}{RGB}{231,248,238}
\definecolor{sgUpHigh}{RGB}{213,240,225}
\definecolor{sgDownLow}{RGB}{253,245,244}
\definecolor{sgDownMid}{RGB}{249,232,231}
\newcommand{\sggainlow}[1]{\cellcolor{sgUpLow}#1}
\newcommand{\sggainmid}[1]{\cellcolor{sgUpMid}#1}
\newcommand{\sggainhigh}[1]{\cellcolor{sgUpHigh}#1}
\newcommand{\sglosslow}[1]{\cellcolor{sgDownLow}#1}
\newcommand{\sglossmid}[1]{\cellcolor{sgDownMid}#1}
\setlength{\tabcolsep}{2.4pt}
\renewcommand{\arraystretch}{1.08}
\resizebox{\textwidth}{!}{
\begin{tabular}{@{}ll*{12}{c}@{}}
\toprule
\rowcolor{sgPanel}
\multicolumn{14}{@{}l}{\textbf{Open-weight models}} \\
\midrule
\multirow{2}{*}{\textbf{Dataset}} & \multirow{2}{*}{\textbf{Split}}
& \multicolumn{3}{c}{\texttt{Gemma-4-26B}}
& \multicolumn{3}{c}{\texttt{Llama-3.1-8B}}
& \multicolumn{3}{c}{\texttt{Mistral-Nemo}}
& \multicolumn{3}{c}{\texttt{Qwen-2.5-7B}} \\
\cmidrule(lr){3-5} \cmidrule(lr){6-8} \cmidrule(lr){9-11} \cmidrule(l){12-14}
& & \textsc{Base} & \textsc{Skill} & $\Delta$ & \textsc{Base} & \textsc{Skill} & $\Delta$ & \textsc{Base} & \textsc{Skill} & $\Delta$ & \textsc{Base} & \textsc{Skill} & $\Delta$ \\
\midrule
\rowcolor{sgStripe}
& IOD    & 78.67\% & 86.00\% & \sggainmid{7.33\%}  & 68.00\% & 70.67\% & \sggainlow{2.67\%}  & 63.33\% & 62.67\% & \sglosslow{-0.67\%} & 64.00\% & 77.33\% & \sggainhigh{13.33\%} \\
\multirow{-2}{*}{\texttt{ALFWorld}} & OOD    & 82.35\% & 93.73\% & \sggainhigh{11.37\%} & 67.45\% & 65.10\% & \sglosslow{-2.35\%} & 58.82\% & 67.06\% & \sggainmid{8.24\%}  & 69.41\% & 82.35\% & \sggainhigh{12.94\%} \\
\rowcolor{sgStripe}
\texttt{LiveCodeBench} & -- & 83.33\% & 83.33\% & 0.00\% & 16.00\% & 20.00\% & \sggainlow{4.00\%} & 11.33\% & 14.00\% & \sggainlow{2.67\%} & 25.33\% & 25.33\% & 0.00\% \\
& All    & 70.00\% & 65.00\% & \sglossmid{-5.00\%} & 17.50\% & 40.00\% & \sggainhigh{22.50\%} & 47.50\% & 50.00\% & \sggainlow{2.50\%} & 40.00\% & 40.00\% & 0.00\% \\
\rowcolor{sgStripe}
\multirow{-2}{*}{\texttt{MCPBench}} & Single & 12.50\% & 12.50\% & 0.00\%  & 0.00\%  & 0.00\%  & 0.00\%  & 0.00\%  & 0.00\%  & 0.00\% & 6.25\%  & 6.25\%  & 0.00\% \\
\texttt{Mind2Web}     & -- & 52.00\% & 53.00\% & \sggainlow{1.00\%}  & 24.00\% & 24.00\% & 0.00\% & 20.00\% & 35.00\% & \sggainhigh{15.00\%} & 27.00\% & 28.00\% & \sggainlow{1.00\%} \\
\rowcolor{sgStripe}
\texttt{PubMedQA}     & -- & 76.00\% & 78.00\% & \sggainlow{2.00\%}  & 62.00\% & 69.00\% & \sggainmid{7.00\%} & 61.00\% & 67.00\% & \sggainmid{6.00\%}  & 66.00\% & 66.00\% & 0.00\% \\
\texttt{ScienceWorld} & -- & 18.00\% & 34.00\% & \sggainhigh{16.00\%} & 33.00\% & 36.00\% & \sggainlow{3.00\%} & 31.00\% & 41.00\% & \sggainhigh{10.00\%} & 23.00\% & 26.00\% & \sggainlow{3.00\%} \\
\rowcolor{sgStripe}
& FTS    & 98.00\% & 98.00\% & 0.00\% & 48.00\% & 48.00\% & 0.00\% & 62.00\% & 66.00\% & \sggainlow{4.00\%} & 54.00\% & 54.00\% & 0.00\% \\
\multirow{-2}{*}{\texttt{SocialMaze}} & UPI    & 14.00\% & 14.00\% & 0.00\% & 16.00\% & 16.00\% & 0.00\% & 16.00\% & 16.00\% & 0.00\% & 12.00\% & 18.00\% & \sggainmid{6.00\%} \\
\midrule
\rowcolor{sgAvg}
\textbf{Avg.} & -- & \textbf{58.49\%} & \textbf{61.76\%} & \textbf{3.27\%} & \textbf{35.20\%} & \textbf{38.88\%} & \textbf{3.68\%} & \textbf{37.10\%} & \textbf{41.87\%} & \textbf{4.77\%} & \textbf{38.70\%} & \textbf{42.33\%} & \textbf{3.63\%} \\
\midrule
\rowcolor{sgPanel}
\multicolumn{14}{@{}l}{\textbf{Proprietary models}} \\
\midrule
\multirow{2}{*}{\textbf{Dataset}} & \multirow{2}{*}{\textbf{Split}}
& \multicolumn{3}{c}{\texttt{Claude-Haiku-4.5}}
& \multicolumn{3}{c}{\texttt{GPT-5.4-Nano}}
& \multicolumn{3}{c}{\texttt{GPT-5.4-Mini}}
& \multicolumn{3}{c}{\texttt{Grok-4-Fast}} \\
\cmidrule(lr){3-5} \cmidrule(lr){6-8} \cmidrule(lr){9-11} \cmidrule(l){12-14}
& & \textsc{Base} & \textsc{Skill} & $\Delta$ & \textsc{Base} & \textsc{Skill} & $\Delta$ & \textsc{Base} & \textsc{Skill} & $\Delta$ & \textsc{Base} & \textsc{Skill} & $\Delta$ \\
\midrule
\rowcolor{sgStripe}
& IOD    & 60.00\% & 66.67\% & \sggainmid{6.67\%}  & 82.67\% & 94.00\% & \sggainhigh{11.33\%} & 89.33\% & 96.67\% & \sggainmid{7.33\%}  & 66.67\% & 77.33\%  & \sggainhigh{10.67\%} \\
\multirow{-2}{*}{\texttt{ALFWorld}} & OOD    & 61.18\% & 64.71\% & \sggainlow{3.53\%}  & 72.94\% & 94.90\% & \sggainhigh{21.96\%} & 93.33\% & 97.25\% & \sggainlow{3.92\%}  & 65.10\% & 80.39\%  & \sggainhigh{15.29\%} \\
\rowcolor{sgStripe}
\texttt{LiveCodeBench} & -- & 56.00\% & 56.00\% & 0.00\% & 59.33\% & 65.33\% & \sggainmid{6.00\%} & 59.33\% & 65.33\% & \sggainmid{6.00\%} & 84.67\% & 84.67\% & 0.00\% \\
& All    & 67.50\% & 82.50\% & \sggainhigh{15.00\%} & 80.00\% & 80.00\% & 0.00\%  & 75.00\% & 75.00\% & 0.00\%  & 47.50\% & 65.00\%  & \sggainhigh{17.50\%} \\
\rowcolor{sgStripe}
\multirow{-2}{*}{\texttt{MCPBench}} & Single & 68.75\% & 81.25\% & \sggainhigh{12.50\%} & 18.75\% & 31.25\% & \sggainhigh{12.50\%} & 43.75\% & 43.75\% & 0.00\%  & 12.50\% & 12.50\%  & 0.00\% \\
\texttt{Mind2Web}     & -- & 60.00\% & 59.00\% & \sglosslow{-1.00\%} & 51.00\% & 57.00\% & \sggainmid{6.00\%}  & 57.00\% & 61.00\% & \sggainlow{4.00\%}  & 67.00\% & 72.00\%  & \sggainmid{5.00\%} \\
\rowcolor{sgStripe}
\texttt{PubMedQA}     & -- & 74.50\% & 74.50\% & 0.00\%  & 67.00\% & 69.00\% & \sggainlow{2.00\%}  & 68.00\% & 72.00\% & \sggainlow{4.00\%}  & 80.00\% & 80.00\%  & 0.00\% \\
\texttt{ScienceWorld} & -- & 41.00\% & 49.00\% & \sggainmid{8.00\%}  & 29.00\% & 54.00\% & \sggainhigh{25.00\%} & 40.00\% & 43.00\% & \sggainlow{3.00\%}  & 43.00\% & 56.00\%  & \sggainhigh{13.00\%} \\
\rowcolor{sgStripe}
& FTS    & 58.00\% & 80.00\% & \sggainhigh{22.00\%} & 64.00\% & 80.00\% & \sggainhigh{16.00\%} & 56.00\% & 80.00\% & \sggainhigh{24.00\%} & 98.00\% & 100.00\% & \sggainlow{2.00\%} \\
\multirow{-2}{*}{\texttt{SocialMaze}} & UPI    & 22.00\% & 20.00\% & \sglosslow{-2.00\%} & 18.00\% & 18.00\% & 0.00\%  & 10.00\% & 14.00\% & \sggainlow{4.00\%}  & 24.00\% & 24.00\%  & 0.00\% \\
\midrule
\rowcolor{sgAvg}
\textbf{Avg.} & -- & \textbf{57.73\%} & \textbf{62.52\%} & \textbf{4.79\%} & \textbf{54.27\%} & \textbf{64.35\%} & \textbf{10.08\%} & \textbf{59.17\%} & \textbf{64.80\%} & \textbf{5.63\%} & \textbf{58.84\%} & \textbf{65.19\%} & \textbf{6.35\%} \\
\bottomrule
\end{tabular}
}
\endgroup
\vspace{-5pt}
\end{table*}

\begin{figure*}[t]
\centering
\includegraphics[width=0.99\textwidth]{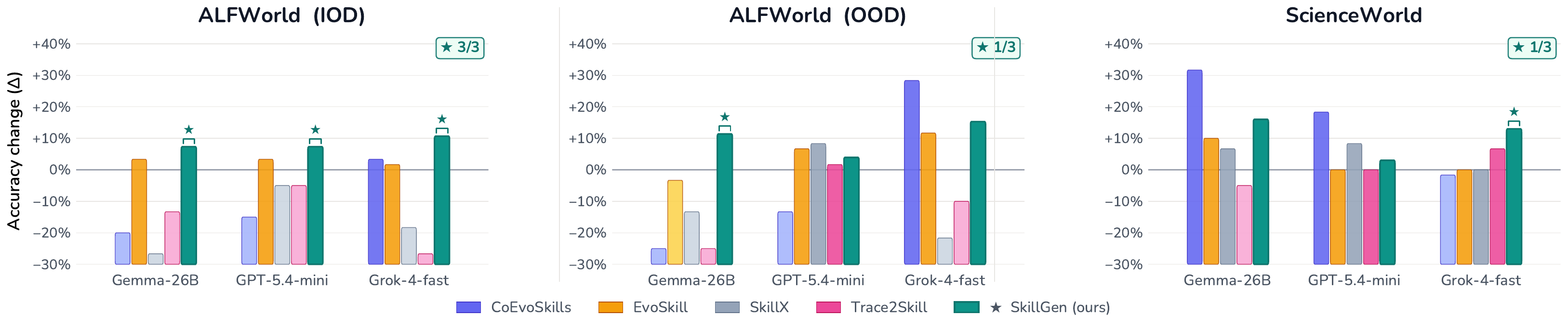}
\vspace{-4pt}
\caption{\textbf{Comparison with skill-generation baselines.} Accuracy improvement ($\Delta$) from adding a generated skill across representative benchmark--model entries. Mini, Grok, and Gemma denote \texttt{GPT-5.4-Mini}, \texttt{Grok-4-Fast}, and \texttt{Gemma-4-26B}, respectively. All methods use the same evaluation harness.}
\label{fig:baselines-comparison}
\vspace{-8pt}
\end{figure*}
Table~\ref{tab:main-results} shows three main patterns: (i) \textsc{SkillGen} improves average accuracy for all eight base agents, with gains from $+3.27$ to $+10.08$ percentage points; (ii) the effect holds across both open-weight models ($+3.27$ to $+4.77$ pp) and proprietary models ($+4.79$ to $+10.08$ pp); and (iii) out of 80 held-out benchmark--split--model entries, 50 improve, 25 remain unchanged, and only 5 show regressions.
The largest gains appear on procedural, multi-step benchmarks: \texttt{ALFWorld} improves in 14 of 16 entries, and \texttt{ScienceWorld} improves for all eight agents.
Further, \textsc{SkillGen} is especially useful when the base model has enough task capability to execute a learned procedure but still has room to improve.

\begin{figure*}[t]
\centering
\includegraphics[width=0.9\textwidth]{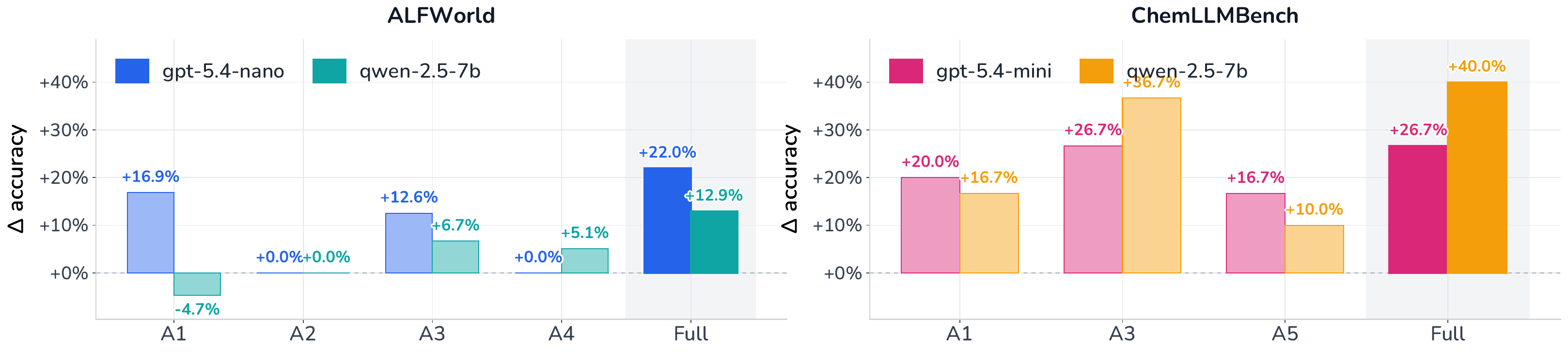}
\vspace{-6pt}
\caption{\textbf{\textsc{SkillGen} ablations.} $\Delta$ accuracy over a shared no-skill baseline on \texttt{ALFWorld} (OOD) and \texttt{ChemLLMBench} yield prediction. \textbf{A1}: ICL ($k=3$) instead of the induced skill; \textbf{A2}: no refinement; \textbf{A3}: no verification gate; \textbf{A4}: no Failure Lessons; \textbf{A5}: plain-text skill (no script+reference bundle); \textbf{Full}: complete \textsc{SkillGen}. \textbf{Full} wins on every dataset--model pair, showing that each component contributes.}
\label{fig:ablation-bars}
\vspace{-10pt}
\end{figure*}

\begin{figure*}[t]
\centering
\includegraphics[width=0.99\textwidth]{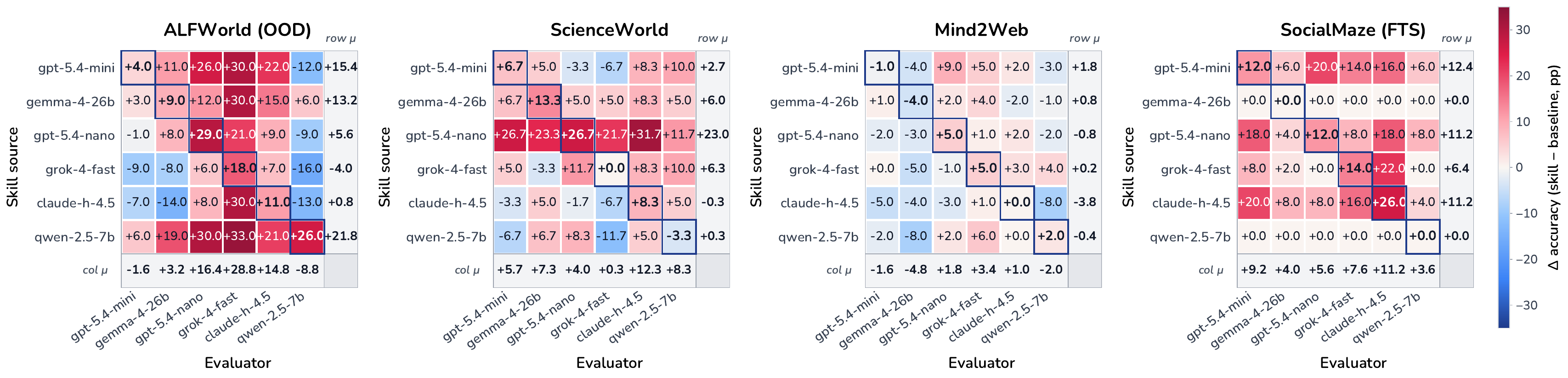}
\vspace{-5pt}
\caption{\textbf{Cross-model skill transferability.} Each heatmap reports $\Delta$ accuracy when a skill generated by a source model (row) is executed by an evaluator model (column). Diagonal cells are self-transfer, while off-diagonal cells are cross-model transfer. Right and bottom margins show transfer-out and transfer-in means, respectively; color saturates at $\pm 30$ pp. The transfer matrix is evaluated on a shared pool of 100 held-out instances per benchmark, distinct from the main evaluation split, to ensure that baseline trajectories are consistent across all 36 (source, evaluator) pairs.}
\label{fig:transfer-matrix}
\vspace{-10pt}
\end{figure*}

\noindent\textbf{\rqtag{rq2}{2} How does \textsc{SkillGen} compare with state-of-the-art automatic skill-generation baselines?}

We compare \textsc{SkillGen} against four recent skill-generation baselines: \textbf{Trace2Skill}~\citep{ni2026trace2skill}, \textbf{SkillX}~\citep{wang2026skillx}, \textbf{EvoSkill}~\citep{alzubi2026evoskill}, and \textbf{CoEvoSkills}, a co-evolutionary baseline instantiated from EvoSkills~\citep{zhang2026evoskills}. The implementation details of baselines are in Appendix~\ref{app:baseline-details}.
We evaluate on \texttt{ALFWorld} IOD, \texttt{ALFWorld} OOD, and \texttt{ScienceWorld}, using three agent models spanning different capability tiers and providers. Figure~\ref{fig:baselines-comparison} summarizes $\Delta$ accuracy across the benchmark--model entries.
We observe that \textsc{SkillGen} leads to consistent gain across settings and achieves the largest overall improvement.

\begin{wrapfigure}{r}{0.45\textwidth}
\vspace{-8pt}
\centering
\includegraphics[width=\linewidth]{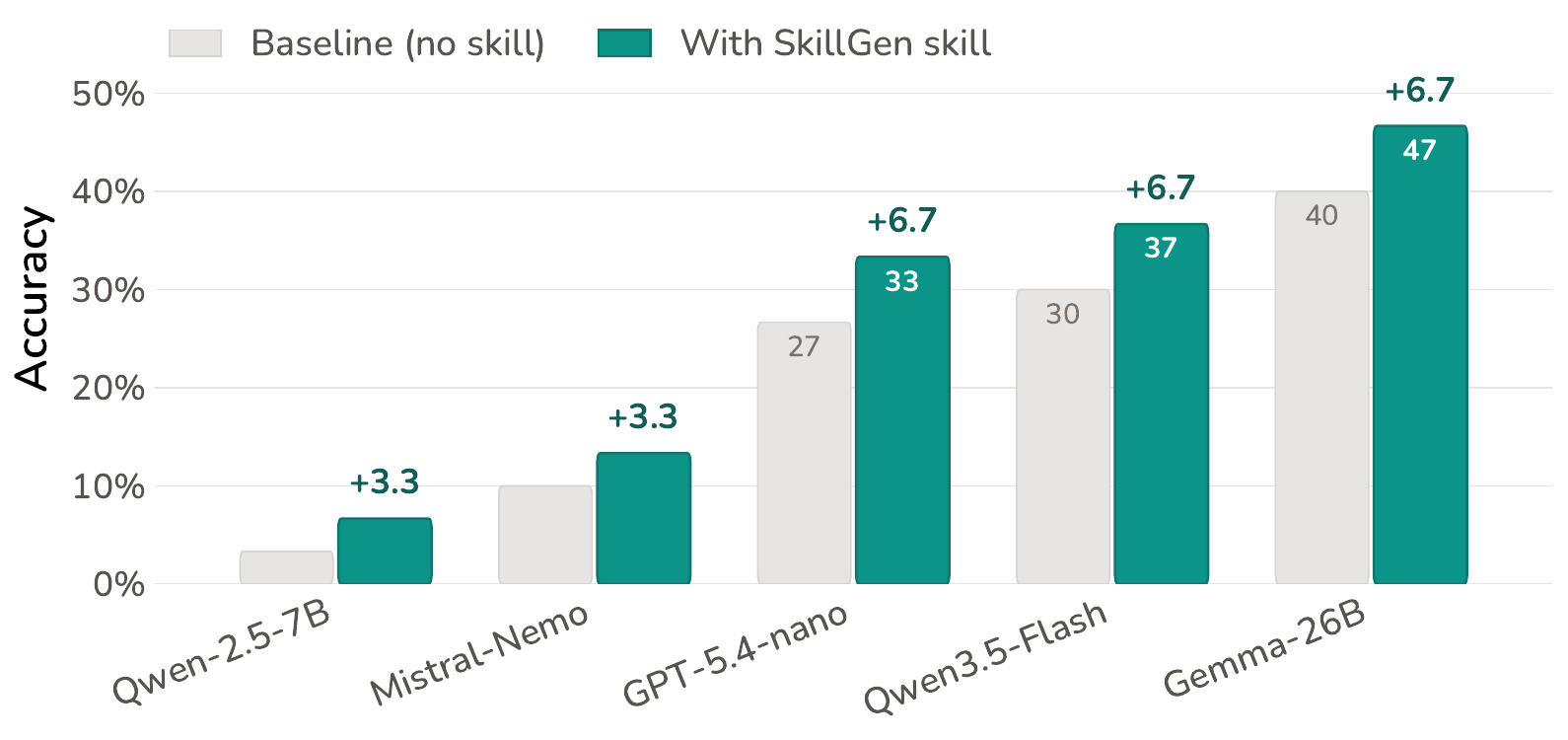}
\vspace{-8pt}
\caption{\textbf{Insights for $\tau$-\texttt{Bench}.} Held-out accuracy on $\tau$-\texttt{Bench} retail for the five models where the \textsc{SkillGen} verification gate activated. Gray bars are no-skill baselines and teal bars apply the induced skill; deltas are absolute percentage-point changes.}
\label{fig:tau-bench-before-after}
\vspace{-15pt}
\end{wrapfigure}

\noindent\textbf{\rqtag{rq3}{3} Which components are necessary for reliable skill construction?}

Figure~\ref{fig:ablation-bars} compares \textsc{SkillGen} against several ablations on representative prediction tasks and shows that each component is relevant for overall performance.
We find: (i)~the induced skill outperforms simple $k=3$ demonstration reuse, so the improvement is not just retrieval; (ii)~refinement and the verification gate are both needed for reliable interactive-task gains, because early candidates can repair some failures while introducing regressions; and (iii)~task-specific skill structure is also relevant (e.g., the failure patterns help on \texttt{ALFWorld} OOD and the script+reference bundle helps on \texttt{ChemLLMBench}).
The complete \textsc{SkillGen} system achieves the best result on every dataset--model pair in the ablation study.

\noindent\textbf{\rqtag{rq4}{4} Are generated skills transferable across agents?}

We evaluate transfer by reusing the final \textsc{SkillGen} skill from one source model without retraining it, then executing it with a different evaluator model on \texttt{ALFWorld} OOD, \texttt{ScienceWorld}, \texttt{Mind2Web}, and \texttt{SocialMaze} FTS.
Each transferred skill is compared against the evaluator's own no-skill baseline; skills marked `deprecated' by the source pipeline are retained as no-op skills. Figure~\ref{fig:transfer-matrix} shows that \textsc{SkillGen} produces skills that often transfer across models, but relevant is the choice of skill-generating model. Across 120 off-diagonal comparisons, 70\% are non-negative, and 42\% exceed $+5$ pp.
We see a clear pattern: transferable skills are not simply written by the strongest baseline agents; on \texttt{ALFWorld}, \texttt{Qwen-2.5-7B} is the best skill-generating model on average, while, on \texttt{ScienceWorld}, \texttt{GPT-5.4-Nano} is best.

\noindent\textbf{\rqtag{rq5}{5} In which additional task regimes does \textsc{SkillGen} provide useful gains?}

Figure~\ref{fig:tau-bench-before-after} evaluates long-horizon retail tool use on $\tau$-\texttt{Bench}.
\textsc{SkillGen} improves every model whose skill passes the verification gate, with an average gain of $+5.3$ pp; models whose candidates fail the verification gate are left unchanged rather than exposed to an unverified intervention.

\begin{figure*}[t]
\centering
\includegraphics[width=0.99\textwidth]{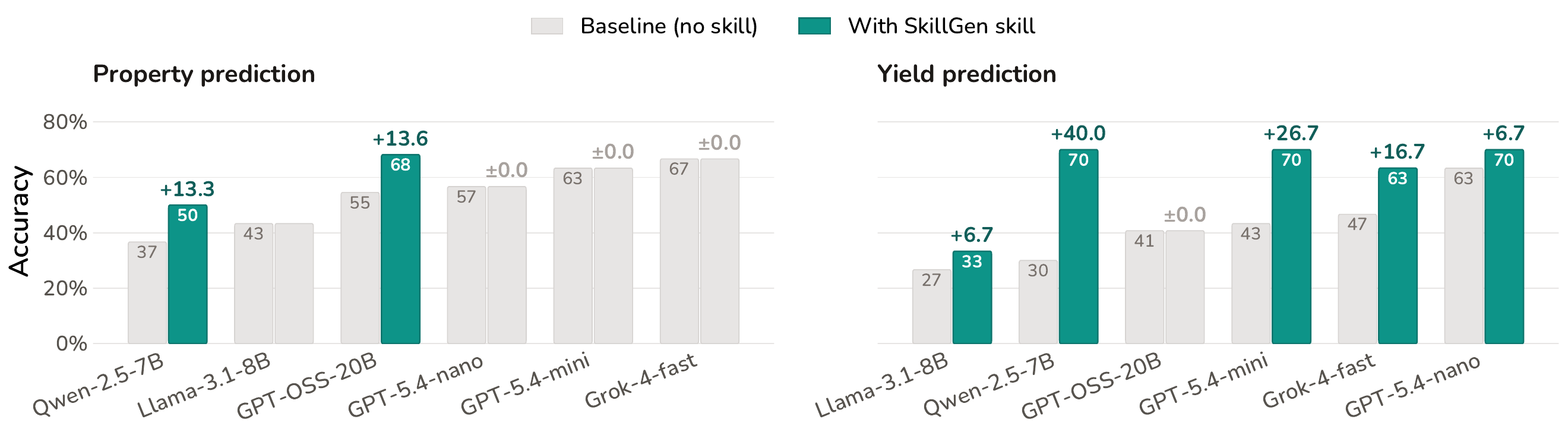}
\vspace{-6pt}
\caption{\textbf{Insights for \texttt{ChemLLMBench}.} Held-out accuracy on \texttt{ChemLLMBench} property prediction (left) and yield prediction (right). Gray bars are no-skill baselines and teal bars apply the \textsc{SkillGen} skill; bars labeled ``$\pm 0.0$'' or ``gate off'' indicate no measurable change or rejection by the verification gate.}
\label{fig:chemllmbench-before-after}
\vspace{-8pt}
\end{figure*}

\begin{figure*}[t]
\centering
\includegraphics[width=0.99\textwidth]{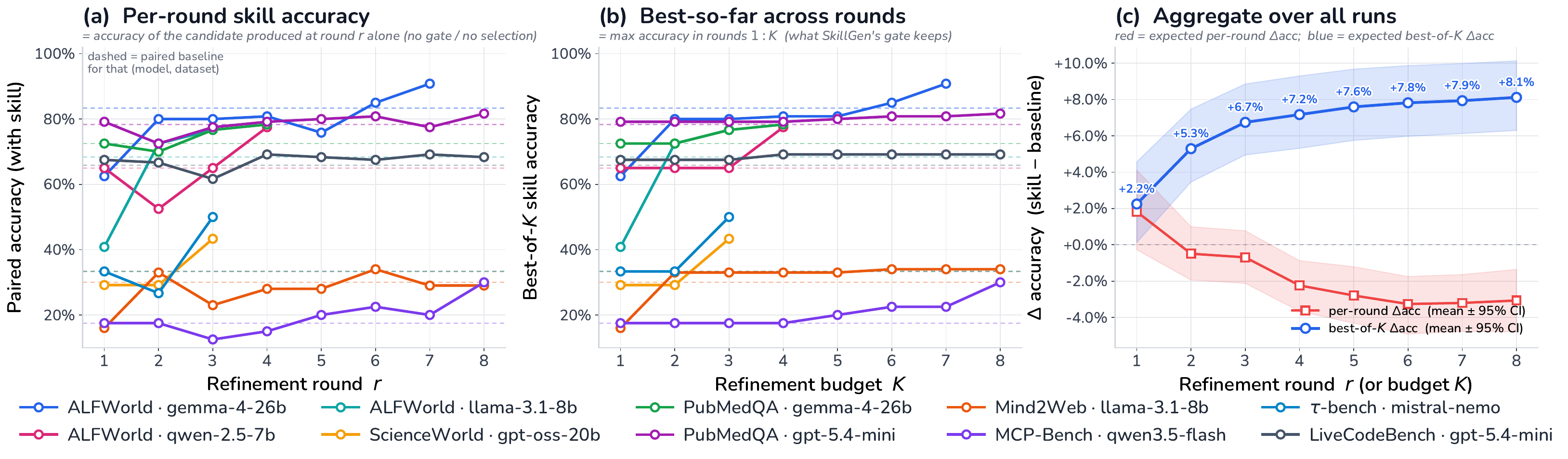}
\caption{\textbf{Refinement rounds vs. skill accuracy.} Each refinement round produces one candidate skill evaluated on the construction-time verification subset. \textbf{(a)} Per-round candidate accuracy for representative runs, with dashed no-skill baselines. \textbf{(b)} Best-so-far accuracy under a budget of $K$ rounds. \textbf{(c)} Aggregate mean $\Delta$ accuracy over all runs with 95\% bootstrap confidence intervals.}
\label{fig:rounds-curve}
\vspace{-10pt}
\end{figure*}

Figure~\ref{fig:chemllmbench-before-after} evaluates script- and reference-augmented skills on \texttt{ChemLLMBench}.
Here, yield prediction benefits, with an average improvement of $+16.1$ pp across six models, while property prediction is more knowledge-bound and improves only for a small subset of agents.
Together, these results suggest that resource-bundle skills are especially useful when the task is procedurally learnable, rather than being primarily a matter of recalling domain facts.

\noindent\textbf{\rqtag{rq6}{6} Why select the best verified refinement round instead of using the latest candidate?}

Figure~\ref{fig:rounds-curve} provides an empirical justification for treating refinement as best-of-$K$ search rather than using the latest candidate.
Per-round candidates are noisy: by round 8, the latest candidate has expected $\Delta=-3.1$ pp, while the best verified candidate reaches $+8.1$ pp (i.e. a gap of $\sim$ 11 pp).
Thus, the verification gate is not just a safety check; it is also the selection mechanism that turns unstable refinement trajectories into a reliable final skill.

\noindent\textbf{\rqtag{rq7}{7} What qualitative failure modes and insights emerge from the generated skills?}

We also inspect manually logged benchmark summaries and per-model skill-analysis reports; a detailed qualitative analysis is in Appendix~\ref{app:failure-analysis}.
Our analysis supports three takeaways: (i) the verification gate removes many harmful candidate skills, although accepted skills can still overgeneralize on held-out instances; (ii) residual failures in interactive environments are usually incomplete procedure execution rather than single local action mistakes; and (iii) chemistry and coding failures often reflect grounding or global-structure limits that a reusable inference-time skill cannot always repair.
These observations align with the design of \textsc{SkillGen}: summary diagnostics for contrastive learning identify recurring procedural gaps, while construction-time verification limits the deployment of harmful skill interventions.

\begin{ack}
This paper is supported by the DAAD program "Konrad Zuse Schools of Excellence in Artificial Intelligence", sponsored by the Federal Ministry of Education and Research.
\end{ack}

\newpage
\bibliographystyle{plainnat}
\bibliography{references}

\newpage
\appendix

\section{Algorithm}
\label{app:algorithm}

Algorithm~\ref{alg:skillgen-loop} summarizes the full \textsc{SkillGen} construction procedure.

\begin{algorithm}[h]
\caption{\textsc{SkillGen}: contrastive induction with generation-verification-refinement loop}
\label{alg:skillgen-loop}
\begin{algorithmic}[1]
\Require induction subset $\mathcal{D}_{\mathrm{ind}}$, construction-time verification subset $\mathcal{D}_{\mathrm{ver}}$, base agent $\mathcal{A}$, evaluator $\mathcal{E}$, round budget $K\ge1$, gate parameters $g_{\mathrm{abs}}\in\mathbb{Z}_{\ge0}$ and $g_{\mathrm{rel}}\in[0,1]$
\Ensure skill $s^{\star}$ marked $\mathrm{active}$ or $\mathrm{deprecated}$
\Statex \emph{Stage \circledblue{1}}\ \textit{Baseline elicitation}
\State Collect induction trajectories $\mathcal{B}=\{(x_i,\tau_i^{0},y_i^{0})\}_{i=1}^{n}$ by rolling out $\mathcal{A}$ on $\mathcal{D}_{\mathrm{ind}}$
\State Cache verification baselines $\mathcal{B}_{\mathrm{ver}}=\{(\tilde{x}_j,\tilde{\tau}^{0}_{j},b_j)\}_{j=1}^{m}$ by rolling out $\mathcal{A}$ on $\mathcal{D}_{\mathrm{ver}}$
\Statex \emph{Stage \circledblue{2}}\ \textit{Contrastive behavioral induction}
\State Induction agent analyzes trajectories into $\mathcal{Z}=(a_0,\mathcal{F},\mathcal{S},\mathcal{C})$
\Statex \emph{Stage \circledblue{3}}\ \textit{Generation--verification--refinement}
\State Initialize $\Phi^{(0)}\gets\varnothing$,\ \ $g^{\star}\gets-\infty$,\ \ $s^{\star}\gets\varnothing$
\State Set gate threshold $\gamma_m\gets\max\{g_{\mathrm{abs}},\lceil g_{\mathrm{rel}}m\rceil,1\}$
\For{$r=1,\ldots,K$}
    \State Generation agent computes $s^{(r)}\gets
           \begin{cases}
               \operatorname{Gen}(\mathcal{Z}), & \text{for } r=1,\\
               \operatorname{Refine}(s^{(r-1)},\mathcal{Z},\Phi^{(r-1)}), & \text{for }  r>1
           \end{cases}$
    \State Verification agent evaluates $s^{(r)}$ on all $\tilde{x}_j\in\mathcal{D}_{\mathrm{ver}}$ and computes $G_m(s^{(r)})$
    \State Verification agent builds feedback $\Phi^{(r)}$ from repairs, regressions, and unresolved failures
    \If{$G_m(s^{(r)})>g^{\star}$}
        \State $g^{\star}\gets G_m(s^{(r)})$,\ \ $s^{\star}\gets s^{(r)}$
    \EndIf
\EndFor
\State Mark $s^{\star}$ as $\mathrm{active}$ if $g^{\star}\ge\gamma_m$, else mark it as $\mathrm{deprecated}$
\State \Return $s^{\star}$
\end{algorithmic}
\end{algorithm}

\newpage

\section{Related Work}
\label{sec:related_work}

\noindent\textbf{Agent skills.}
Early LLM agents augmented models with external tool use~\citep{schick2023toolformer,qin2024toolllm} or tool creation~\citep{qian2023creator}, while interaction-based agents such as ReAct~\citep{yao2023react}, Reflexion~\citep{shinn2023reflexion}, ExpeL~\citep{zhao2024expel}, and Voyager~\citep{wang2023voyager} showed that trajectories can support reusable reasoning and action routines.

Agent skills provide a first-class abstraction that often follows the Anthropic Agent Skills standard~\citep{zhang2025agentskills,anthropic2025skills}, which defines skills as composable bundles of instructions, scripts, and resources loaded dynamically at inference time.
Recent surveys systematize skill architecture, acquisition, security, and deployment~\citep{xu2026agentskillsurvey,jiang2026sok}.
On the construction side, Trace2Skill~\citep{ni2026trace2skill}, EvoSkill~\citep{alzubi2026evoskill}, EvoSkills~\citep{zhang2026evoskills}, and SkillX~\citep{wang2026skillx} synthesize skills from agent experience, while SkillWeaver~\citep{zheng2025skillweaver}, WebXSkill~\citep{wang2026webxskill}, AutoSkill~\citep{yang2026autoskill}, and SkillRL~\citep{xia2026skillrl} study web, dialogue, and RL deployment regimes.
\textsc{SkillGen} differs by making failure analysis central: it clusters error patterns, identifies capability boundaries, and verifies induced skills through multi-agent collaboration with a construction-time deployment rule over paired repairs and regressions.

\noindent\textbf{Synthetic data for LLM agents.}
Self-Instruct~\citep{wang2023selfinstruct} developed LLM-bootstrapped instruction generation, which inspiring later extension such as Alpaca~\citep{taori2023alpaca}, WizardLM~\citep{xu2024wizardlm}, and further quality improvements through reasoning traces, curation, synthetic textbooks, and taxonomy-driven generation~\citep{mukherjee2023orca,zhou2023lima,gunasekar2023textbooks,li2024glan}.
Recent pipelines scale synthetic data with agentic flows or self-synthesis~\citep{mitra2024agentinstruct,xu2025magpie}, controllable generation and verification~\citep{huang2025datagen}, domain-specific tool-aware construction~\citep{huang2025chemorch}, verified visual question answering~\citep{bao2024autobench} and risk-injected safety trajectories~\citep{huang2025guardrail}.

For agent-specific capabilities, AgentTuning~\citep{zeng2024agenttuning}, Agent-FLAN~\citep{chen2024agentflan}, FireAct~\citep{chen2023fireact}, and Instruct-SkillMix~\citep{kaur2025instructskillmix} construct trajectory or skill-composition data for instruction tuning.
These methods produce data for \emph{model fine-tuning}; in contrast, \textsc{SkillGen} synthesizes \emph{inference-time skills}---structured bundles loaded without parameter updates---that target capability gaps exposed by systematic failure analysis.

\newpage

\section{Experimental Details}
\label{app:experimental-details}

\subsection{Model Details}
\label{app:model-details}

\begin{table}[h]
\centering
\caption{Base agent models used across the reported experiments. Open-weight indicates that model weights are publicly available; proprietary models are accessed through hosted APIs.}
\label{tab:model-details}
\begingroup
\small
\definecolor{sgAppHeader}{RGB}{246,248,251}
\definecolor{sgAppStripe}{RGB}{250,251,253}
\newcommand{\sgOpenYes}{\textcolor{green!55!black}{\(\checkmark\)}}
\newcommand{\sgOpenNo}{\textcolor{red!65!black}{\(\times\)}}
\setlength{\tabcolsep}{7pt}
\renewcommand{\arraystretch}{1.08}
\begin{tabular}{@{}llc@{}}
\toprule
\rowcolor{sgAppHeader}
\textbf{Model} & \textbf{Provider} & \textbf{Open-weight} \\
\midrule
\rowcolor{sgAppStripe}
\texttt{Gemma-4-26B} & Google & \sgOpenYes \\
\texttt{Llama-3.1-8B} & Meta & \sgOpenYes \\
\rowcolor{sgAppStripe}
\texttt{Mistral-Nemo} & Mistral AI & \sgOpenYes \\
\texttt{Qwen-2.5-7B} & Alibaba Cloud & \sgOpenYes \\
\rowcolor{sgAppStripe}
\texttt{GPT-OSS-20B} & OpenAI & \sgOpenYes \\
\texttt{Qwen3.5-Flash} & Alibaba Cloud & \sgOpenNo \\
\rowcolor{sgAppStripe}
\texttt{Claude-Haiku-4.5} & Anthropic & \sgOpenNo \\
\texttt{GPT-5.4-Nano} & OpenAI & \sgOpenNo \\
\rowcolor{sgAppStripe}
\texttt{GPT-5.4-Mini} & OpenAI & \sgOpenNo \\
\texttt{Grok-4-Fast} & xAI & \sgOpenNo \\
\rowcolor{sgAppStripe}
\texttt{Nemotron-120B} & NVIDIA & \sgOpenNo \\
\bottomrule
\end{tabular}
\endgroup
\vspace{-5pt}
\end{table}

\subsection{Datasets and Splits}
\label{app:datasets-splits}

All reported accuracy changes are based on the following comparative assessment: for each held-out test instance, we roll out the same base agent once without a skill and once with the generated skill, using the same instance identifier and random seed. Unless otherwise specified, we use seed 42 and keep the skill-training dataset and held-out test pool disjoint. Within the skill-training dataset, \textsc{SkillGen} uses an induction subset for trajectory analysis and a construction-time verification subset for refinement and selection, matching the usual train/validation/test separation. Table~\ref{tab:appendix-splits} summarizes the controlled split protocol for the benchmark-specific studies that require explicit sampling.

\begin{table}[h]
\centering
\caption{Controlled split protocol for benchmark-specific studies.}
\label{tab:appendix-splits}
\begingroup
\small
\definecolor{sgDetailHeader}{RGB}{246,248,251}
\definecolor{sgDetailStripe}{RGB}{250,251,253}
\setlength{\tabcolsep}{4pt}
\renewcommand{\arraystretch}{1.08}
\begin{tabular}{@{}p{0.18\textwidth}p{0.24\textwidth}rrp{0.30\textwidth}@{}}
\toprule
\rowcolor{sgDetailHeader}
\textbf{Benchmark} & \textbf{Held-out test split} & \textbf{Constr.} & \textbf{Test} & \textbf{Notes} \\
\midrule
\rowcolor{sgDetailStripe}
$\tau$-\texttt{Bench} & Retail, task-disjoint & 30 & 30 & User simulator: \texttt{GPT-5.4-Nano} \\
\texttt{ALFWorld} & IOD (\texttt{valid\_seen}) & 500 & 150 & In-distribution; task types overlap with train \\
\rowcolor{sgDetailStripe}
\texttt{ALFWorld} & OOD (\texttt{valid\_unseen}) & 500 & 255 & Out-of-distribution; novel task types \\
\texttt{ChemLLMBench} & Property / yield prediction & 30 & 30 & Subtasks scored independently \\
\rowcolor{sgDetailStripe}
\texttt{LiveCodeBench} & \texttt{test\_release\_v6} & 50 & 150 & Sampled from release v6; seed 42 \\
\texttt{MCPBench} & All tools & 64 & 40 & Multi-tool coverage \\
\rowcolor{sgDetailStripe}
\texttt{MCPBench} & Single tool & 40 & 16 & Single-tool coverage \\
\texttt{Mind2Web} & Task-level disjoint & 100 & 100 & Web navigation; task-level split \\
\rowcolor{sgDetailStripe}
\texttt{PubMedQA} & Sampled & 150 & 100 & Biomedical QA; seed 42 \\
\texttt{ScienceWorld} & \texttt{dev} & 150 & 100 & Long-horizon multi-step tasks \\
\rowcolor{sgDetailStripe}
\texttt{SocialMaze} & FTS & 60 & 50 & Fixed-target social navigation \\
\texttt{SocialMaze} & UPI & 60 & 50 & Unpredictable persona interaction \\
\bottomrule
\end{tabular}
\endgroup
\vspace{-5pt}
\end{table}

\subsection{Model Routing and Auxiliary Roles}
\label{app:model-routing}

Each baseline or skill-augmented rollout is executed by the base agent model listed in Table~\ref{tab:model-details}; the auxiliary models used inside \textsc{SkillGen} never replace the base agent during rollout. To isolate the effect of the generated skill from the capability of the skill-writing model, we use a fixed auxiliary model, \texttt{GPT-5.4-Mini}, for the induction agent, generation agent, and verification agent across all base agents. Non-OpenAI model calls are routed through OpenRouter. Embeddings for clustering and skill-card merging use \texttt{text-embedding-3-small}. Decoding is deterministic with temperature 0; the default output budget is 4{,}096 tokens, increased to 16{,}384 tokens for skill generation.

\subsection{\textsc{SkillGen} Hyperparameters}
\label{app:skillgen-hyperparameters}

Unless otherwise noted, all runs use the same benchmark-specific configuration template. The induction stage uses at most eight failure clusters and eight success clusters, with adaptive $k$-means clustering over $k\in[2,8]$ and a target cluster size of 15. The contrastive module keeps up to 20 nearest failure--success pairs. The generation prompt receives up to six failure clusters, six success clusters, and eight contrastive observations; web search is disabled.

The main experiments use a maximum refinement budget of eight rounds. For candidate verification, the verification gate evaluates uniformly sampled construction-time verification instances from the skill training dataset, using a 70/30 induction/verification split with at least four verification instances when the pool is small. The deployment decision follows the rule in \S\ref{sec:paired-verification}: a skill is accepted only if its construction-time verification net gain $G_m(s)$ satisfies $G_m(s)\ge \max\{2,\lceil 0.05m\rceil,1\}$.
We additionally run up to 30 baseline-success guard checks to expose regressions on already-solved instances. Skills that fail the gate are persisted with status \texttt{deprecated}; downstream evaluation treats them as empty interventions, so cells labeled ``gate off'' report zero change rather than an unverified skill. The pipeline uses four workers for independent runs, and the verification agent's feedback stage uses eight workers.

\subsection{Token Cost Analysis}
\label{app:token-cost}

\begin{table}[h]
\centering
\caption{Token cost of \textsc{SkillGen}. \emph{Train} is the one-time construction budget; \emph{\textsc{Base}} and \emph{\textsc{Skill}} are average tokens per call.}
\label{tab:token-analysis}
\vspace{3pt}
\begingroup
\footnotesize
\definecolor{sgTokHeader}{RGB}{246,248,251}
\definecolor{sgTokStripe}{RGB}{250,251,253}
\definecolor{sgTokAvg}{RGB}{255,253,244}
\setlength{\tabcolsep}{3.8pt}
\renewcommand{\arraystretch}{1.04}
\begin{tabular}{@{}lrrr@{}}
\toprule
\rowcolor{sgTokHeader}
\textbf{Benchmark} & \textbf{Train} & \textbf{\textsc{Base}} & \textbf{\textsc{Skill}} \\
\rowcolor{sgTokHeader}
                   & \textbf{(M tok)} & \textbf{(tok/call)} & \textbf{(tok/call)} \\
\midrule
\rowcolor{sgTokStripe}
\texttt{ScienceWorld} & 2.2  & 1{,}630 & 1{,}977 \\
\texttt{PubMedQA}     & 2.7  & 1{,}173 & 2{,}429 \\
\rowcolor{sgTokStripe}
\texttt{Mind2Web}     & 5.2  & 4{,}482 & 5{,}919 \\
\texttt{MCPBench}     & 7.5  & 4{,}847 & 6{,}000 \\
\rowcolor{sgTokStripe}
$\tau$-\texttt{Bench} & 10.2 & 5{,}813 & 6{,}358 \\
\midrule
\rowcolor{sgTokAvg}
\textbf{Mean}   & \textbf{5.6} & \textbf{3{,}589} & \textbf{4{,}537} \\
\rowcolor{sgTokAvg}
\textbf{Median} & \textbf{5.2} & \textbf{4{,}482} & \textbf{5{,}919} \\
\bottomrule
\end{tabular}
\endgroup
\vspace{-5pt}
\end{table}

Table~\ref{tab:token-analysis} separates one-time construction cost from per-call inference overhead. All values are computed per model and then averaged within each benchmark. The skill-construction pipeline, including baseline trajectory collection, induction, generation, refinement, and verification, is a one-time cost per model--benchmark pair, ranging from 2.2M tokens on \texttt{ScienceWorld} to 10.2M on $\tau$-\texttt{Bench} (mean 5.6M). Using \texttt{GPT-5.4-Mini} standard API prices (\(\$0.75\)/M input tokens and \(\$4.50\)/M output tokens),\footnote{\url{https://openai.com/api/pricing}, accessed April 2026.} and the prompt/output mix observed in our training logs, the mean construction budget corresponds to approximately \(\$8.2\) per generated skill. This cost is paid once for a model--benchmark pair, after which the same skill can be reused across subsequent rollouts and repeated evaluations. The per-call columns show that retrieval keeps inference prompts in the same few-thousand-token regime: the median skill-augmented call is 5{,}919 tokens, and the largest absolute per-call average in the table is 6{,}358 tokens on $\tau$-\texttt{Bench}.

\paragraph{Compute resources.}
All experiments are orchestrated locally but executed through hosted LLM APIs routed through OpenRouter. The base-agent and auxiliary-model routing is reported in Appendix~\ref{app:model-routing}; token budgets are reported in Table~\ref{tab:token-analysis}; and concurrency settings are reported in Appendix~\ref{app:skillgen-hyperparameters}. Because model inference is served by third-party API providers, the underlying accelerator type, memory configuration, and provider-side scheduling are not exposed to us. We therefore report reproducible API-level compute usage in terms of model calls, token budgets, and worker concurrency. Local compute is used only for orchestration, logging, clustering, and evaluation bookkeeping.

\subsection{Skill-Generation Baselines}
\label{app:baseline-details}

\paragraph{Trace2Skill.}
For Trace2Skill~\citep{ni2026trace2skill}, we run one no-skill rollout over the training pool. Success-branch and error-branch analysts process trajectories in parallel, and their proposed patches are consolidated through hierarchical LLM merging. We preserve the original prompt structure and do not impose an additional output schema beyond the shared Markdown skill wrapper.

\paragraph{SkillX.}
For SkillX~\citep{wang2026skillx}, we run two refinement rounds. Each round rolls out the current library, extracts skill cards from successful trajectories, clusters cards by cosine similarity at threshold 0.80 using \texttt{text-embedding-3-small}, merges clusters with an LLM, and filters cards with an LLM quality score threshold of 3/5. The retained library is capped at 12 cards before being canonicalized into the final skill.

\paragraph{EvoSkill.}
For EvoSkill~\citep{alzubi2026evoskill}, we maintain a frontier of $k=3$ candidate programs for four iterations. A proposer chooses among \texttt{add\_new}, \texttt{edit}, and \texttt{keep} operations based on failures from a fixed validation subset, and a builder emits the next candidate library. Admission requires the new candidate to outperform the weakest frontier member on the same validation subset.

\paragraph{CoEvoSkills.}
We instantiate the co-evolutionary baseline from EvoSkills~\citep{zhang2026evoskills} using an information-isolated surrogate verifier. The surrogate writes binary natural-language assertions, judges rollouts against those assertions, and returns structured diagnostics to the skill generator. To match the rollout budget of the other baselines, we use three outer iterations and two inner verifier iterations. Because the shared evaluation interface consumes Markdown-formatted skills rather than executable multi-file bundles, the surrogate assertions are LLM-judged rather than executed as code.

\paragraph{Scope of the baseline comparison.}
Our baseline comparison evaluates all methods under the same deployment problem studied in this paper: synthesize one fixed, auditable inference-time skill for each benchmark--model pair, and evaluate that fixed intervention on held-out instances. This requires adapting methods that natively construct or retrieve from multi-skill libraries, such as SkillX and EvoSkill, into the same single-skill interface used by \textsc{SkillGen}. We emphasize that this controlled adaptation is not intended to measure the full native deployment potential of those systems under all possible library sizes, retrieval policies, or routing strategies. Instead, it supports a like-for-like comparison of skill synthesis quality when the deployed artifact must be one fixed skill.

This design also reduces evaluation instability and selection bias. If library-based methods were allowed to select different skills at test time, held-out performance would conflate skill construction with additional design choices such as library size, retrieval scoring, context-budget allocation, routing policy, and stochastic skill selection. Those choices are important in their own right, but they introduce extra degrees of freedom that are not shared by all methods and can make a comparative evaluation less stable (and potentially unfair). The resulting comparison should therefore be interpreted as a controlled single-skill adaptation of each baseline, rather than as a claim that the adapted baselines exhaust their native multi-skill capabilities. All reproduced baselines are adapted to this shared single-skill evaluation interface: each method emits one Markdown-formatted skill per benchmark--model pair, which is then injected and evaluated by the shared paired rollout harness. The base agent model is used for task rollouts, while \texttt{GPT-5.4-Mini} is used for auxiliary extraction, merging, proposal, and judging steps.

\paragraph{No-tool comparison protocol.}
For the skill-generation baseline comparison, we use only benchmark settings in which the evaluated skill does not rely on external tools, generated helper scripts, or reference-resource loading. To ensure a like-for-like comparison, \textsc{SkillGen}'s optional script and reference components are disabled in these runs: every method, including \textsc{SkillGen}, emits a single Markdown-formatted natural-language skill, i.e., $s=(u,a,\varnothing,\varnothing)$. No method is allowed to provide executable helper functions, generated tools, retrieval documents, or calls to \texttt{skill\_load\_reference}. All skills are injected through the same prompt slot and evaluated with the same paired rollout harness. Thus, the comparison isolates the quality of the synthesized natural-language skill rather than differences in tool availability.

\subsection{Evaluation Metrics and Gate-Off Handling}
\label{app:evaluation-metrics}

For every evaluated benchmark--split--model cell, we report baseline accuracy, skill-augmented accuracy, and the paired difference $\Delta=\mathrm{acc}_{\mathrm{skill}}-\mathrm{acc}_{\mathrm{base}}$ in percentage points. We also record repair counts (baseline wrong, skill correct), regression counts (baseline correct, skill wrong), and net gain as repair minus regression. When \textsc{SkillGen} marks a skill as \texttt{deprecated} during construction-time verification, evaluation reuses the no-skill baseline for the skill side. This convention makes rejected skills explicit and prevents an unverified prompt from introducing hidden regressions.

\subsection{t-SNE Visualizations}
\label{app:tsne}
\begin{figure*}[t]
\centering
\includegraphics[width=0.8\textwidth]{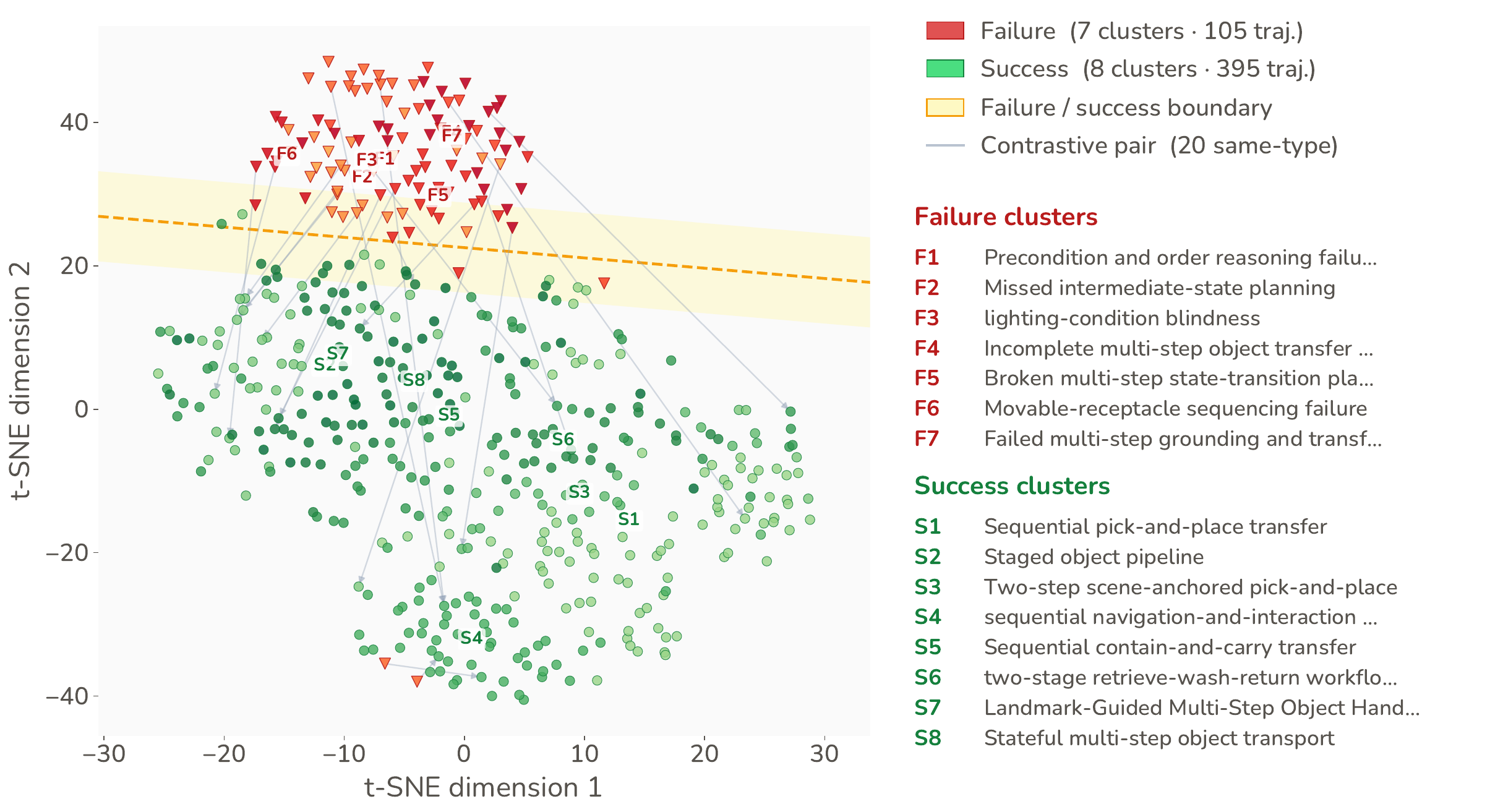}

\caption{t-SNE visualization of SkillGen's induction on ALFWorld (gpt-5.4-nano). Red triangles (F1–F7) are failure trajectories; green circles (S1–S8) are successes. Gray arrows link each failure to its nearest same-type success (20 contrastive pairs). The yellow band marks the decision boundary between the two populations. Failures cluster compactly in the upper region (recurring planning errors), while successes spread broadly (diverse solving strategies), motivating the contrastive analysis that drives skill generation.}
\label{fig:tsne}

\end{figure*}

Fig.~\ref{fig:tsne} shows the t-SNE visualization of the contrastive induction of SkillGen on ALFWorld (gpt-5.4-nano).

\subsection{Failure Analysis}
\label{app:failure-analysis}

We complement the aggregate gains in Section~\ref{sec:experiments} with a qualitative inspection of the archived benchmark summaries and per-model skill-analysis reports. Rather than listing isolated examples, we distill four recurring findings about when skill augmentation still fails.

\paragraph{\textbf{The verification gate substantially reduces harm, but accepted skills can still regress on held-out instances.}}
The clearest evidence is the contrast between rejected and accepted negative skills. In LiveCodeBench, rejected skills would have reduced \texttt{Qwen-2.5-7B} from 25.33\% to 16.00\% and \texttt{Gemma-4-26B} from 83.33\% to 80.67\%; these cells are therefore reported as zero-delta after gating. Similar rejected regressions appear in Mind2Web and PubMedQA. However, filtering is not perfect. On ALFWorld OOD, an accepted skill for \texttt{Llama-3.1-8B} still lowers accuracy from 67.45\% to 65.10\%, with 51 repairs but 57 regressions, and on ChemLLMBench yield prediction \texttt{Mistral-Nemo} drops from 43.33\% to 20.00\%, with only three repairs against ten regressions. The failure mode here is overgeneralization: a skill that looks beneficial on the verification subset can still perturb correct baseline behavior on the final split.

\paragraph{\textbf{In interactive environments, the dominant residual error is incomplete procedure execution rather than local action selection.}}
Both ALFWorld and ScienceWorld exhibit this pattern. For \texttt{Llama-3.1-8B} on ALFWorld, the training-time analysis records all failures, and the largest cluster, with 65 cases, is labeled \emph{incomplete dependency planning}. These trajectories often begin the first plausible subgoal but omit a later prerequisite, such as turning on a lamp before inspection or using an intermediate receptacle before transport. ScienceWorld shows the same phenomenon at the level of experimental procedure. For \texttt{Gemma-4-26B}, the analysis contains 110 failures out of 150 construction instances, with large clusters labeled \emph{ungrounded action planning} (25), \emph{incomplete task-sequence planning} (20), and \emph{incomplete goal-to-action planning} (19). In both benchmarks, the agent usually recognizes the task theme, but fails to maintain the full ordered procedure needed to finish it.

\paragraph{\textbf{On chemistry tasks, failures are driven less by missing facts than by incorrect grounding of reaction roles and decision criteria.}}
For \texttt{Qwen-2.5-7B} on ChemLLMBench yield prediction, the training analysis reports 27 failures out of 30 examples, with two dominant clusters: \emph{superficial reaction-feasibility assessment} (14) and \emph{reaction-role misparsing in cross-coupling} (13). A typical failure is that the model recognizes a familiar reaction family, but confuses substrates with catalysts, bases, solvents, or additives, and then predicts yield from a generic template rather than checking the actual electrophile, nucleophile, ligand/base combination, and substrate scope. This helps explain why resource-bundle skills are especially effective on yield prediction: the gain comes from enforcing a grounded checking procedure, not from merely restating chemical knowledge.

\paragraph{\textbf{On code-generation tasks, skill augmentation cannot fully compensate for missing global problem structure.}}
LiveCodeBench failures for \texttt{Qwen-2.5-7B} are dominated by upstream reasoning errors rather than surface-level implementation mistakes. The training analysis reports 113 failures out of 150 problems, with major clusters labeled \emph{incomplete algorithmic modeling} (19), \emph{incomplete algorithm realization} (14), and \emph{structure-mapping failure} (6). Typical trajectories either pursue brute-force search where an invariant or transformation is required, apply a local greedy heuristic where dynamic programming is needed, or emit truncated code after only partially deriving the solution. This suggests that a general skill can regularize recurring reasoning habits, but it cannot reliably rescue cases where the model never identifies the right combinatorial structure in the first place.

\section{Broader Impacts}
\label{sec:broader-impacts}

\textsc{SkillGen} aims to make LLM agents more reliable and easier to adapt without retraining, which could reduce the manual effort required to build task-specific agent skills and make procedural knowledge more inspectable through human-readable skill artifacts. This may benefit scientific, coding, web, and tool-use workflows where reusable guidance and verification can improve consistency. At the same time, any method that improves agent performance can also improve agents used for harmful or unintended purposes, and generated skills may overgeneralize, introduce regressions on unseen cases, or amplify mistakes in domains where incorrect tool use has real consequences. The risks are especially relevant when skills are deployed in open-ended environments, safety-sensitive applications, or workflows involving external tools and resources. \textsc{SkillGen}'s paired verification provides a partial check against these harms by making regressions visible during construction. However, these checks are limited to the evaluated task distribution, so they should be paired with application-specific safety evaluation, human review of generated skills, access controls, and ongoing monitoring before deployment.

\end{document}